\documentclass[]{ceurart}

\usepackage{amsmath}
\usepackage{amssymb}
\usepackage{amsfonts}
\usepackage{graphicx}

\usepackage{multirow}

\usepackage{caption}
\usepackage{subcaption}

\usepackage[export]{adjustbox}

\begin{document}

\copyrightyear{2023}
\copyrightclause{Copyright for this paper by its authors.
  Use permitted under Creative Commons License Attribution 4.0
  International (CC BY 4.0).}

\conference{CLEF 2023: Conference and Labs of the Evaluation Forum, September 18–21, 2023, Thessaloniki, Greece}

\title{Strategies to Harness the Transformers' Potential: UNSL at eRisk 2023}


\author[1,2]{Horacio Thompson}[%
email=hjthompson@unsl.edu.ar,
]

\address[1]{Universidad Nacional de San Luis (UNSL), Ejército de Los Andes 950, San Luis, C.P. 5700, Argentina}

\address[2]{Consejo Nacional de Investigaciones Científicas y Técnicas (CONICET), San Luis, Argentina}

\author[1, 2]{Leticia Cagnina}[%
email=lcagnina@unsl.edu.ar,
]

\author[1]{Marcelo Errecalde}[%
email=merreca@unsl.edu.ar,
]

\begin{abstract}
The CLEF eRisk Laboratory explores solutions to different tasks related to risk detection on the Internet. In the 2023 edition, Task 1 consisted of searching for symptoms of depression, the objective of which was to extract user writings according to their relevance to the BDI Questionnaire symptoms. Task 2 was related to the problem of early detection of pathological gambling risks, where the participants had to detect users at risk as quickly as possible. Finally, Task 3 consisted of estimating the severity levels of signs of eating disorders. \\ 
Our research group participated in the first two tasks, proposing solutions based on \emph{Transformers}. For Task 1, we applied different approaches that can be interesting in information retrieval tasks. Two proposals were based on the similarity of \emph{contextualized embedding vectors}, and the other one was based on \emph{prompting}, an attractive current technique of machine learning. For Task 2, we proposed three fine-tuned models followed by decision policy according to criteria defined by an early detection framework. One model presented extended vocabulary with important words to the addressed domain. In the last task, we obtained good performances considering the decision-based metrics, ranking-based metrics, and runtime. In this work, we explore different ways to deploy the predictive potential of \emph{Transformers} in eRisk tasks.
\end{abstract}


\maketitle

\section{Introduction}
The Early Risk Prediction on the Internet (eRisk) laboratory proposes solving different challenges to explore evaluation methodologies, effectiveness metrics, and practical applications for risk detection in social networks. Through its editions \cite{losada2016erisk, losada2017erisk, losada2018overview, losada2019overview, losada2020erisk, parapar2021overview, parapar2022overview, parapar2023erisk}, several tasks have been proposed on different domains, promoting the participating teams to propose innovative solutions that solve the tasks in the best possible way. Our research group has actively participated in eRisk editions with notable contributions \cite{errecalde2017temporal, funez2018unsl, burdisso2019unsl, loyola2021unsl, loyola2022unsl}.
In the 2023 edition \cite{parapar2023erisk}, a new task was introduced: Task 1, which involved searching for symptoms of depression in a collection of user writings. Task 2 was a continuation of the 2022 edition of the problem on early risk detection of pathological gambling. Finally, Task 3 consisted of estimating the severity level of signs of eating disorders.

The neural architectures known as \emph{Transformers} proposed by Vaswani et al. \cite{vaswani2017attention} have caused a true revolution in the artificial intelligence field. Numerous studies have shown the performance of \emph{Transformers} to solve a wide variety of natural language processing tasks, with models such as BERT \cite{devlin2018bert}, GPT-2 \cite{radford2019language}, and GPT-3 \cite{brown2020language}. Motivated by the relevant role of \emph{Transformers}, we have proposed to address this year's tasks by applying approaches based on this architecture.

Our research group participated in Tasks 1 and 2, focusing on strategies that take advantage of the predictive power of \emph{Transformers}. For Task 1, we presented three proposals. The first two were based on measuring the similarity of embeddings extracted from pre-trained models. We represented the writings of the collection and the symptoms of the BDI Questionnaire using verbs, adjectives, and nouns. From these terms, we obtained word embeddings using a BERT model adjusted for depression tasks, which allowed us to measure the closeness between the writings and the symptoms considering the task domain and then obtain the final rankings. The third solution consisted of applying one of the most attractive approaches currently known as \emph{prompting}, which takes advantage of the predictive power of a pre-trained language model, adapting it to a particular task \cite{liu2023pre}.
We applied the \emph{Fixed-prompt LM Tuning} technique \cite{li2021prefix} by fitting the RoBERTa model \cite{liu2019roberta} on samples containing pre-defined prompts. Since labeled data were unavailable, we created a dataset of 200 samples per symptom using \emph{ChatGPT} \cite{openai2021}. We created prompts by concatenating each sample to a simple template with slots to fill. Then, we tuned the RoBERTa model to solve the missing word on these prompts, continuing its previous training. We evaluated the texts of the collection by predicting the prompts, and we associated the responses of the model with the probability of belonging to each one of the symptoms. Finally, we improved the ranking of two of the 21 symptoms with a multiclass classifier obtained by \emph{fine-tuning} on the created dataset. For Task 2, we used an early detection framework \cite{loyola2018learning}, which led us to remarkable results in previous editions \cite{loyola2021unsl, loyola2022unsl}.
The method defines that, to solve an early detection problem, it is necessary to consider two components: one dedicated to solving a user classification problem (classification with partial information or CPI), and the other involves a decision policy to decide when to stop evaluating a user (deciding the moment of classification or DMC). On this occasion, we presented three proposals, for which we applied the BERT model with some variants (CPI component) and defined a decision policy based on the history of predictions that a model performs during user evaluation (DMC component).

Our main contribution in this edition was the application of different strategies based on \emph{Transformers}. We applied novel techniques that may be of interest to works related to information retrieval. For the early detection problem, combining fine-tuned models with a decision policy based on a historic rule allowed us to maximize performance and obtain good results.

\section{Task 1: Search for symptoms of depression}
\label{sec:task1}

From a collection of user posts, the task consisted of building rankings of relevant posts considering the 21 depression symptoms of the BDI Questionnaire. A \emph{relevant sentence} for a symptom S was defined as the user-generated text that provides information about their condition related to S.
According to official data, the collection contains 3,807,115 sentences extracted from 3,107 users. 
Teams could submit up to five proposals, each with 21 rankings of up to 1,000 writings in descending order by score. 
Our research group put forward three solutions: two based on the similarity of contextualized embeddings and one employing the \emph{prompting} technique.

\subsection{Similarity-based proposals}

The strategies consisted of measuring the closeness of the writings to each of the symptoms. 

\vspace{0.1cm} \noindent 
\textbf{Sentences filter.} Due to the large number of samples in the collection, a filter was initially applied to reduce the search space. For this, we analyzed the sentiment of the texts with the VADER processor \cite{hutto2014vader}, and we selected those whose negative polarity score was greater than 0. This fact allowed us to reduce the search space from almost 4 million writings to 1 million, which meant that 25\% had some indication of negativity. 

\vspace{0.1cm} \noindent 
\textbf{Preprocessing steps.} Characters were converted to lowercase, while Unicode and HTML codes were transformed into their corresponding symbols. Web pages and numbers were replaced by the \emph{weblink} and \emph{number} tokens, respectively. Repeated words and spaces were also removed, and emojis were manually replaced with text representations related to the symptoms.

\vspace{0.1cm} \noindent 
\textbf{Symptoms and writings representation.} One way to understand a text is to direct the analysis toward important words that describe its semantics \cite{lin2017structured}. To represent each symptom, ten verbs, ten adjectives, and ten nouns were chosen, which were manually selected based on the information provided in the BDI Questionnaire. To represent writings, we used the \emph{Spacy} parser to extract verbs, adjectives, and nouns. \emph{Stopwords} were discarded, except those words that were present in the symptoms. The next step was to represent the symptoms and writings through \emph{word embeddings} considering the context of the addressed domain. 
We employed a transformer-based approach instead of non-contextualized methods such as \emph{Word2Vec} \cite{mikolov2013efficient} and \emph{FastText} \cite{bojanowski2017enriching}. Thus, a context was defined for each word, and a language model was used to extract its embedding vector. For instance, to represent \emph{feeling} in the context of the symptom \emph{sadness}, it was defined as \emph{feeling is linked to the symptom sadness}. Subsequently, embeddings were obtained using the last layer of a BERT model tuned on samples from users with depression\footnote{Available in: \url{https://huggingface.co/BitanBiswas/depression-detection-bert}.}.

\vspace{0.1cm} \noindent 
\textbf{Embeddings similarity.} The verbs, adjectives, and nouns of the texts were compared with the verbs, adjectives, and nouns of the symptoms using \emph{cosine similarity}, obtaining a table of scores as follows:
\begin{table}[h]
\centering
\begin{tabular}{ l }
$\forall$ \text{ text T, symptom  S: } \\
\hspace{1cm} $\forall$ \text{ V$_T$ in verbs(T), V$_S$ in verbs(S):} \\
\hspace{2cm} \text{context\_V$_T$ = ``V$_T$ \emph{is linked to the symptom} S''} \\
\hspace{2cm} \text{emb\_V$_T$ = extract\_embedding(BERT, context\_V$_T$)} \\
\hspace{2cm} \text{context\_V$_S$ = ``V$_S$ \emph{is linked to the symptom} S''} \\
\hspace{2cm} \text{emb\_V$_S$ = extract\_embedding(BERT, context\_V$_S$)} \\
\hspace{2cm} \text{similarity = similarity(emb\_V$_T$, emb\_V$_S$)} \\
\hspace{1cm} Repeat for adjectives and nouns \\
\end{tabular}
\end{table}
\ \\

\vspace{0.1cm} \noindent 
\textbf{Summary of scores.} To summarize the values of a text according to symptoms, two strategies were considered:

\begin{itemize}
\item 
\emph{\textbf{Similarity-MAX:}} The score of each text is summarized considering the verb, adjective, and noun with the maximum similarity, that is, those closest to the symptom, and then the three values are averaged.
\item
\emph{\textbf{Similarity-AVG:}} The score of each text is summarized considering the verb, adjective, and noun with the best average, and then the values obtained are averaged. Unlike the previous one, the word that, on average, was closest to the symptom is taken to reach the final score.
\end{itemize}

\begin{figure}[htbp]
  \centering
  \begin{subfigure}{0.4\textwidth}
    \includegraphics[width=\linewidth]{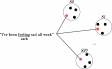}
    \caption{Similarity-MAX}
    \label{fig:sim1a}
  \end{subfigure}
  \hfill
  \begin{subfigure}{0.4\textwidth}
    \includegraphics[width=\linewidth]{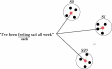}
    \caption{Similarity-AVG}
    \label{fig:sim1b}
  \end{subfigure}
  \caption{Graphic scheme of similarity-based proposals. The distance of a text is observed considering its verb and the verbs of each one of the symptoms (S0, S1,..., S20) for (a) \emph{Similarity-MAX} and (b) \emph{Similarity-AVG}.}
  \label{fig:sim1}
\end{figure}

\begin{figure}[htbp]
\includegraphics[width=\textwidth]{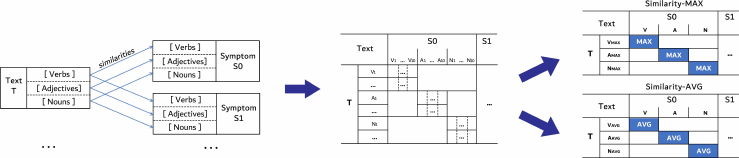}
\caption{Tabular scheme of the similarity-based proposals implementation. The similarities between text T and the symptoms are saved in a scores table and summarized according to \emph{Similarity-MAX} and \emph{Similarity-AVG}.}
\label{fig:sim2}
\end{figure}

\vspace{0.1cm} \noindent 
\textbf{Final rankings.} For both strategies, the final ranking was created by ordering each symptom by score and extracting the first 1,000 writings.
Figures \ref{fig:sim1} and \ref{fig:sim2} show the graphic scheme and the implementation in a tabular format of similarity-based proposals.
\subsection{Prompting-based proposals}

\emph{Prompting} aims to reformulate the original task as a masked language problem and take advantage of the predictive models' ability to complete the missing words of an input text. The method is divided into two steps: 1) \emph{create prompts}, where a text fragment with slots to fill (\emph{template}) is added to each original sentence that the model must complete; 2) \emph{derive the final answer}, where the prediction of the model is mapped to an adequate answer for the original task. Finding suitable prompts can be challenging as it influences the solution's success. It can be solved by manually searching for the best prompt \cite{petroni2019language} or by automatic techniques such as \emph{soft-prompting} \cite{qin2021learning, lester2021power}. On the other hand, \emph{Fixed-prompt LM Tuning} is based on fixing a prompt and continuing the training of the language model to improve its predictions. 
In a \emph{few-shot} scenario, large language models are generally used due to the lack of labeled samples but can only be accessed via an API interface. Because of this, and according to Task 1, we decided to use a small and adjustable model based on \emph{Transformers}, applying the \emph{prompting} paradigm through the \emph{Fixed-prompt LM Tuning} technique.

\vspace{0.1cm} \noindent 
\textbf{Sentences filter and preprocessing steps.} As in the similarity-based proposals, writings that presented negative scores were selected to reduce the search space with the VADER processor, and the same preprocessing was also applied.

\vspace{0.1cm} \noindent 
\textbf{Dataset creation.} 
We created a dataset of 200 samples per symptom to continue training the language model. An application to automatically interact with ChatGPT was implemented as follows: \\
\emph{Query: Generate a list of user texts linked to S. For reference, use words from this list: [w$_1$, w$_2$, ...]} \\
\emph{Answer: Text$_1$, Text$_2$, ... , Text$_{200}$}. \\
Consequently, for each symptom S, we obtained labeled samples \emph{<Text$_1$, S>, <Text$_2$, S>, ... , <Text$_{200}$, S>}. The $w_i$ words were extracted from the terms used to represent the symptoms in the similarity-based proposals.

\vspace{0.1cm} \noindent 
\textbf{Language Model Tuning.} The RoBERTa pre-trained model was imported and tuned based on the masked language problem as follows:
\begin{table}[h]
\centering
\begin{tabular}{ l }
$\forall$ \text{ sample <T, S>} \\
\hspace{1cm} \text{label = T + \emph{``This is linked to S''}} \\
\hspace{1cm} \text{prompt = T + \emph{``This is linked to MASK''}} \\
\hspace{1cm} \text{pred = RoBERTa(prompt)} \\
\hspace{1cm} \text{loss = CrossEntropy(pred, label)} \\
\hspace{1cm} \text{...} \\
\end{tabular}
\end{table}
\ \\
In this way, the model learned to complete missing words, considering the predefined prompt and the samples associated with the symptoms, directing the model's answers toward the task to be solved. For instance, for the sample \emph{<``I've been feeling sad all week'', sadness}>, the prompt ``\emph{I've been feeling sad all week. This is linked to MASK}'' is formed; then, the model should predict that the best word for \emph{MASK} is \emph{sadness}. 

\vspace{0.2cm} \noindent
\textbf{Evaluation of the writings.} The texts of the collection were evaluated using the \emph{prompting} scheme. We defined the prompt using the same template with which the model was tuned. We also created a words dictionary (\emph{verbalizer}) for mapping model predictions to the final output, including the words used to create the dataset. In this way, the tuned model was used to predict the probability that the writings were linked to each symptom.

\newpage \noindent
\textbf{Preliminary ranking.} A table of results was obtained with the first 1,000 writings ordered by probability in decreasing order for each symptom.

\vspace{0.2cm} \noindent
When inspecting the writings obtained in the rankings, we observed that samples not related to depression were found in two of the 21 symptoms: \emph{worthlessness} and \emph{loss of energy}. For the former, finance and economic crisis were the main topics, while for the latter, they were renewable energy, and oil, among others. On the other hand, the ranking of \emph{indecisiveness} and \emph{fatigue} had potential writings of \emph{worthlessness} and \emph{loss of energy}, respectively. Because the probabilities are correlated, the possibility of belonging to one symptom affects the rest. Therefore, we improved the quality of these two symptoms with an extension of the proposal, re-evaluating the writings using a classification model.

\vspace{0.2cm} \noindent
\textbf{Fine-tuning RoBERTa for \emph{worthlessness} y \emph{loss-energy}.} A multiclass classification problem was formulated considering the labels: \emph{worthlessness}, \emph{loss-energy}, and \emph{others}. The training set was defined as follows: for \emph{worthlessness} and \emph{loss-energy}, we used samples extracted from the previously created dataset; for \emph{others}, the remaining symptoms were used, excluding sentences with the \emph{indecisiveness} and \emph{fatigue} labels. Besides, we added texts related to monetary value loss. The \emph{others} class attracted those samples that should not be part of \emph{worthlessness} and \emph{loss-energy}. Then, \emph{fine-tuning} was applied to the RoBERTa model on this training set.

\vspace{0.2cm} \noindent
\textbf{Classification of the writings.} The texts were evaluated by the classification model, recording the probability of each prediction in a new table. The ranking for \emph{worthlessness} and \emph{loss-energy} was created, and we selected the first 1,000 sentences.

\vspace{0.2cm} \noindent
\textbf{Final ranking (\emph{Prompting-Classifier}).} The final proposal consisted of 19 symptom rankings using \emph{prompting} and 2 rankings using the \emph{classifier} explained above.

\subsection{Results}
For evaluating the teams, a writing pool for each symptom was built by selecting the first 50 sentences from all proposals (in total, 37). The organizers carried out a labeling process with three assessors that manually chose which writings were relevant \cite{parapar2023erisk}, resulting in two evaluation schemes: \emph{majority voting} (if 2/3 agreed) and \emph{unanimity} (if 3/3 agreed). Table \ref{tab:rel-sents} shows an extract of the result of the labeling process.

\begin{table}[h]
\caption{Extract from the count of relevant sentences for each symptom of the BDI Questionnaire. \emph{Original}: a writing pool with the first 50 sentences extracted from proposals of all teams. \emph{Majority voting}: 2 of 3 assessors agreed. \emph{Unanimity}: 3 of 3 agreed.}
\label{tab:rel-sents}
\begin{tabular}{lccc}
\hline
\textbf{Symptom}           & \textbf{Original} & \textbf{Majority voting} & \textbf{Unanimity} \\ \hline
Sadness                     & 1110              & 318                      & 179                \\
Pessimism                   & 1150              & 325                      & 104                \\
Past Failure                & 973               & 300                      & 160                \\
Loss of Pleasure            & 1013              & 204                      & 97                 \\
Guilty Feelings             & 829               & 143                      & 83                 \\
Punishment Feelings         & 1079              & 50                       & 21                 \\
...                         & ...               & ...                      & ...                \\ \hline
\end{tabular}
\end{table}

\begin{figure}[t]
    \centering
    \begin{subfigure}[b]{\textwidth}
        \centering
        \includegraphics[width=0.7\textwidth]{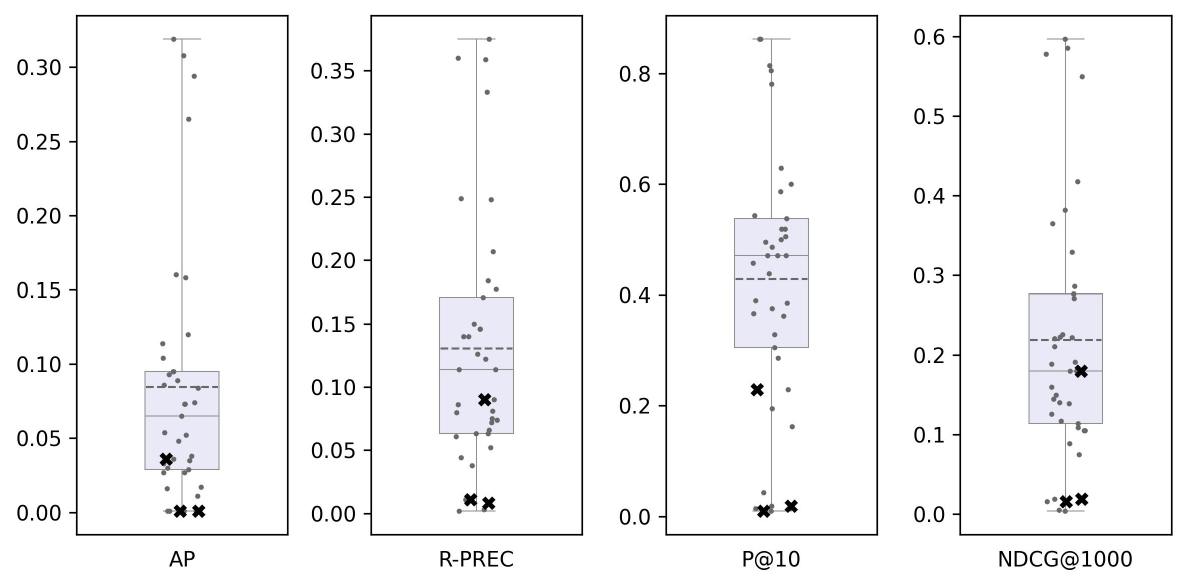}
        \caption{Majority voting}
        \label{fig:bp_majority}
    \end{subfigure}
    
    \begin{subfigure}[htbp]{\textwidth}
        \centering
        \includegraphics[width=0.7\textwidth]{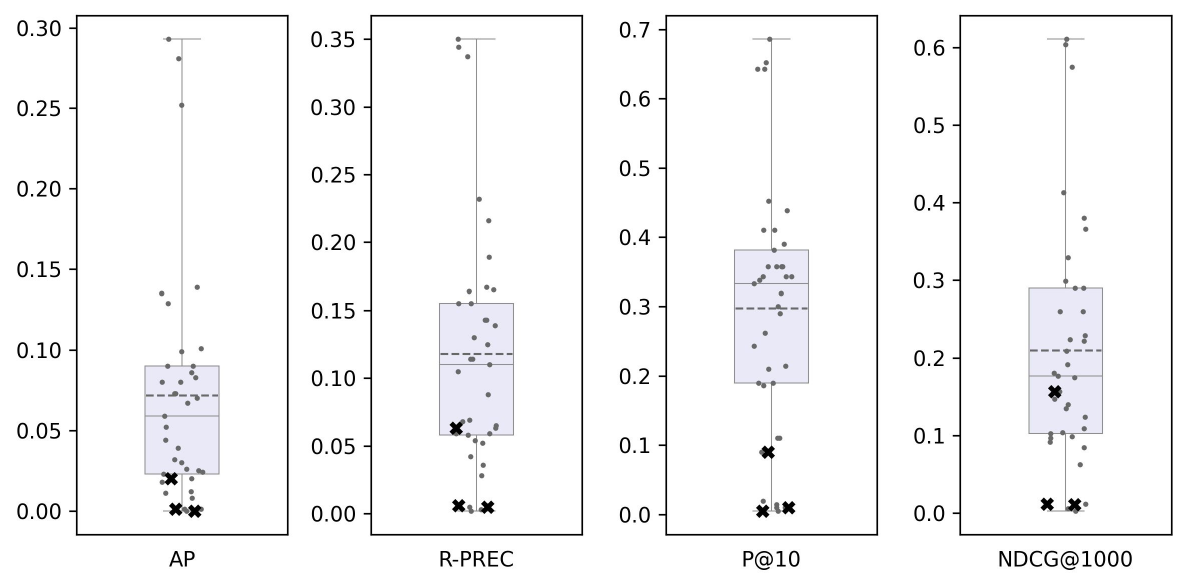}
        \caption{Unanimity}
        \label{fig:bp_unanimity}
    \end{subfigure}
    
    \caption{Performance distribution of the proposals presented by all teams for Task 1, considering (a) \emph{Majority voting} and (b) \emph{Unanimity}. The performances of our team are shown with X marks. The \emph{mean} (dashed line) is included.}
    \label{fig:dist-T1}
\end{figure}

Tables \ref{tab:results-t1a} and \ref{tab:results-t1b} show the results obtained, considering the \emph{majority voting} and \emph{unanimity} schemes. The best results were achieved by Formula-ML. 
The mean, median, and performance distribution of all proposals (Figure \ref{fig:dist-T1}) show that most teams maintained a considerably lower performance than Formula-ML. \emph{Prompting-Classifier} stands out among our proposals with a performance close to that of most teams, considering the AP, R-PREC, and NDCG@1000 metrics.

\begin{table}[htbp]
\caption{Ranking-based evaluation for Task 1. Results are reported according to the metrics Average Precision (AP), R-Precision (R-PREC), Precision at 10 (P@10), and NDCG at 1000 (NDCG@1000), for the majority voting scheme.}
\label{tab:results-t1a}
\begin{tabular}{llcccc}
\hline
\multirow{2}{*}{\textbf{Team}} & \multirow{2}{*}{\textbf{Run}} & \multicolumn{4}{c}{\textbf{Majority voting}}                           \\
                               &                               & \textbf{AP}    & \textbf{R-PREC} & \textbf{P@10}  & \textbf{NDCG@1000} \\ \hline
UNSL                           & Prompting-Classifier          & 0.036          & 0.090           & 0.229          & 0.180              \\
UNSL                           & Similarity-AVG                & 0.001          & 0.008           & 0.010          & 0.016              \\
UNSL                           & Similarity-MAX                & 0.001          & 0.011           & 0.019          & 0.019              \\ \hline
Formula-ML                     & SentenceTransformers\_0.25    & \textbf{0.319} & \textbf{0.375}  & \textbf{0.861} & \textbf{0.596}     \\
Formula-ML                     & SentenceTransformers\_0.1     & 0.308          & 0.359           & \textbf{0.861} & 0.584              \\ \hline
\multicolumn{2}{c}{\textit{Mean}}                              & 0.084          & 0.131           & 0.428          & 0.219              \\
\multicolumn{2}{c}{\textit{Median}}                            & 0.065          & 0.114           & 0.471          & 0.180              \\ \hline
\end{tabular}
\end{table}

\begin{table}[htbp]
\caption{Ranking-based evaluation for Task 1 according to the unanimity scheme.}
\label{tab:results-t1b}
\begin{tabular}{llcccc}
\hline
\multirow{2}{*}{\textbf{Team}} & \multirow{2}{*}{\textbf{Run}} & \multicolumn{4}{c}{\textbf{Unanimity}}                                 \\
                               &                               & \textbf{AP}    & \textbf{R-PREC} & \textbf{P@10}  & \textbf{NDCG@1000} \\ \hline
UNSL                           & Prompting-Classifier          & 0.020          & 0.063           & 0.090          & 0.157              \\
UNSL                           & Similarity-AVG                & 0.000          & 0.005           & 0.005          & 0.011              \\
UNSL                           & Similarity-MAX                & 0.001          & 0.006           & 0.010          & 0.012              \\ \hline
Formula-ML                     & SentenceTransformers\_0.25    & 0.268          & \textbf{0.360}  & \textbf{0.709} & \textbf{0.615}     \\
Formula-ML                     & SentenceTransformers\_0.1     & \textbf{0.293} & 0.350           & 0.685          & 0.611              \\ \hline
\multicolumn{2}{c}{\textit{Mean}}                              & 0.072          & 0.118           & 0.297          & 0.210              \\
\multicolumn{2}{c}{\textit{Median}}                            & 0.059          & 0.110           & 0.333          & 0.177              \\ \hline
\end{tabular}
\end{table}

Several factors may have influenced the general performance of the models, such as the labeling process or the way of evaluating them. As observed in Table \ref{tab:rel-sents}, the number of relevant sentences was considerably less than the original number of samples extracted from all the proposals, and, in turn, distant values are observed between \emph{majority voting} and \emph{unanimity} for each symptom. This fact could indicate that the assessors had different opinions when interpreting sentences, probably due to the concept of \emph{relevance}. On the other hand, it would be important to consider the performance of the models at the symptom level since there were significant differences in the number of relevant sentences between symptoms.

\section{Task 2: Early Detection of Signs of Pathological Gambling}
\label{sec:T1}

The goal was to detect, as early as possible, users that showed signs of pathological gambling. The challenge was divided into two stages: a \emph{training stage}, where the participants experimented with data extracted from previous editions, and a \emph{test stage}, where a client application interacted with a server, defining an \emph{early} environment. This last process was divided into rounds in which the client requested the next post of users and, according to the number of predictive models, evaluated them and returned a response to the server.

Early risk detection can be analyzed as a multi-objective problem, where the challenge is to find an adequate balance between the precision in identifying risky users and the minimum time required for that decision to be reliable. In \cite{loyola2022unsl}, our research group applied the early classification framework \cite{loyola2018learning} by using a BERT model with extended vocabulary (CPI component) and a decision policy based on a historic rule (DMC component). In this edition, we presented three proposals, improving the CPI and DMC components to maximize the performance of the final models.

\subsection{Datasets}

Table \ref{tab:corpus_T2} shows the detail of the corpora available to solve the task. The eRisk2021 and eRisk2022 corpora were used to train the models, as well as UNSL2021\_train and UNSL2021\_valid created in \cite{loyola2021unsl}. The eRisk2023 corpus was used for the organizers to evaluate the participating models. It contains 4.7\% of positive users compared to 7\% and 3.9\% of eRisk2021 and eRisk2022, respectively. Furthermore, the number of words per post was considerably higher than in previous editions, which may be a relevant factor for the models' performance when evaluating longer posts. On the other hand, the UNSL2021\_train and UNSL2022\_valid corpora have considerably fewer posts than the eRisk corpora.

\begin{table}[htbp]
\caption{Details of the corpora used for Task 2. The corpora of the different eRisk editions are shown, as well as the corpora created by our team in previous editions. The number of users (total, positives, and negatives) and the number of posts of each corpus are reported. The median, minimum, and maximum number of posts per user and words per post in each corpus are detailed.}
\label{tab:corpus_T2}
\begin{tabular}{|l|ccc|c|ccc|ccc|}
\hline
\multicolumn{1}{|c|}{\multirow{2}{*}{\textbf{Corpus}}} & \multicolumn{3}{c|}{\textbf{\#users}}                                                  & \multirow{2}{*}{\textbf{\#posts}} & \multicolumn{3}{c|}{\textbf{\#posts per user}}                                       & \multicolumn{3}{c|}{\textbf{\#words per post}}                                       \\ \cline{2-4} \cline{6-11} 
\multicolumn{1}{|c|}{}                                 & \multicolumn{1}{c|}{\textbf{Total}} & \multicolumn{1}{c|}{\textbf{Pos}} & \textbf{Neg} &                                   & \multicolumn{1}{c|}{\textbf{Med}} & \multicolumn{1}{c|}{\textbf{Min}} & \textbf{Max} & \multicolumn{1}{c|}{\textbf{Med}} & \multicolumn{1}{c|}{\textbf{Min}} & \textbf{Max} \\ \hline
eRisk2023                                              & \multicolumn{1}{c|}{2,174}          & \multicolumn{1}{c|}{103}          & 2071         & 1,102,871                         & \multicolumn{1}{c|}{327.33}       & \multicolumn{1}{c|}{10}           & 2,004        & \multicolumn{1}{c|}{28.9}         & \multicolumn{1}{c|}{1}            & 12,779       \\
eRisk2022                                              & \multicolumn{1}{c|}{2,079}          & \multicolumn{1}{c|}{81}           & 1,998        & 1,177,590                         & \multicolumn{1}{c|}{297}          & \multicolumn{1}{c|}{3}            & 2,001        & \multicolumn{1}{c|}{11}           & \multicolumn{1}{c|}{0}            & 6,728        \\
eRisk2021                                              & \multicolumn{1}{c|}{2,348}          & \multicolumn{1}{c|}{164}          & 2,184        & 1,130,799                         & \multicolumn{1}{c|}{244}          & \multicolumn{1}{c|}{10}           & 2,001        & \multicolumn{1}{c|}{11}           & \multicolumn{1}{c|}{1}            & 8,241        \\
UNSL2021\_train                                        & \multicolumn{1}{c|}{1,746}          & \multicolumn{1}{c|}{286}          & 1,460        & 158,924                           & \multicolumn{1}{c|}{51}           & \multicolumn{1}{c|}{31}           & 1,188        & \multicolumn{1}{c|}{20}           & \multicolumn{1}{c|}{1}            & 7,479        \\
UNSL2021\_valid                                        & \multicolumn{1}{c|}{1,746}          & \multicolumn{1}{c|}{286}          & 1,460        & 161,204                           & \multicolumn{1}{c|}{53}           & \multicolumn{1}{c|}{31}           & 1,337        & \multicolumn{1}{c|}{20}           & \multicolumn{1}{c|}{1}            & 3,234        \\ \hline
\end{tabular}
\end{table}

\subsection{CPI components: Models}

Each model was trained and validated by combining different corpora with an 85/15 split. We used the BERT model applying the \emph{fine-tuning} process to adjust it to the classification task. A limitation of the BERT architecture is that it only supports 512 input tokens. Thus, we improved the posts extracted from each user by selecting those posts with some indication of negativity by the VADER processor, and the first 512 tokens were taken. We also used a scheduler to automatically adjust the Learning Rate during \emph{fine-tuning}, improving the convergence and performance of the model. Finally, the best model for each proposal was chosen considering the F1 metric over the positive class (F$_1$+).

\vspace{0.1cm} \noindent 
\textbf{UNSL\#0: Classic BERT model} \\
\emph{Training set.} Combination of the eRisk2021, UNSL2021\_train, and UNSL2021\_valid corpora. Extraction of posts with some indication of negativity. \\
\emph{Preprocessing steps.} Characters were converted to lowercase, while Unicode and HTML codes were transformed into their corresponding symbols. Web pages and numbers were replaced by the \emph{weblink} and \emph{number} tokens, respectively. Repeated words and spaces were also removed. \\
\emph{Hyperparameters for fine-tuning.} Architecture = `BERT-based-uncased', optimizer = `AdamW', LR = 3E-5, scheduler = `LinearSchedulerWarmup', batch\_size = 8, and n\_epochs = 3. 

\vspace{0.1cm} \noindent 
\textbf{UNSL\#1: BERT model with an extended vocabulary} \\
We extended the BERT vocabulary using important words to the addressed domain extracted from an external model. The SS3 model \cite{burdisso2019text} was trained to classify users on the available corpora, and we selected the first 40 words according to the confidence values on the positive class. \\
\emph{Training set, preprocessing steps, and hyperparameters for fine-tuning.} The same as the UNSL\#0 model.

\vspace{0.1cm} \noindent 
\textbf{UNSL\#2: Classic BERT model on all available data} \\
Considering the same hyperparameters of the UNSL\#0 model, the \emph{fine-tuning} process was applied using all available data (eRisk2021, eRisk2022, UNSL2021\_train, and UNSL2021\_valid).

\subsection{DMC component: Decision Policy}
The best decision policy for the models described above was evaluated using a mock server\footnote{Available in: \url{https://github.com/jmloyola/erisk\_mock\_server}.}. This tool simulates the eRisk challenge through the rounds of posts and answers submissions, and then it calculates the final results according to the decision and ranking-based metrics. It was useful since the performance of CPI models can drastically change when evaluated in an \emph{early} environment.
A client application was defined to manage the interaction with the server. When it receives a round of posts, the system preprocesses the writings, invokes the predictive models (CPI), and applies a decision policy (DMC). To take advantage of the 512 input tokens that the BERT architecture admits, the application uses the last N=10 posts (posts window), linking the current post with previous posts. With the mock server, the client application, and the predictive models, different decision policies were evaluated using the F$_{1}$+ and F$_{latency}$ metrics.

\vspace{0.1cm} \noindent 
\textbf{Decision policy based on a historic rule} \\
The historic rule defines that if the current prediction and the last \emph{M} predictions exceed a \emph{threshold} (limit probability to predict a positive user), the client application must issue a risky user alarm; otherwise, it is necessary to continue the user evaluation. In addition, the rule has the \emph{min\_delay} parameter, which defines the moment when it will start to apply. We obtained that the best parameters were \emph{threshold} = 0.7, \emph{M} = 10, and \emph{min\_delay} = 10. \\
\emph{Performance in an early environment.} The models were evaluated using the eRisk2022 corpus. UNSL\#0: F$_{1}$+ = 0.88 and F$_{latency}$ = 0.83; UNSL\#1: F$_{1}$+ = 0.84 and F$_{latency}$ = 0.81. UNSL\#2 was not tested as it was trained with all the corpora, including eRisk2022.
    
\subsection{Results}
Table \ref{tab:decision_T2} shows the results obtained by our team according to the decision-based metrics. The best results were achieved by ELiRF-UPV\#0, as well as other proposals had good results. Considering the average level among all the teams, our models achieved remarkable results in the F$_{1}$, ERDE$_{50}$, and F$_{latency}$ metrics. The UNSL\#1 and UNSL\#2 models showed similar performance, outperforming UNSL\#0 on the same metrics. Regarding ERDE$_{5}$, the three models obtained the same performance as the mean among all teams.

\begin{table}[htbp]
\caption{Decision-based evaluation results for Task 2. The best team taking into account the F$_{1}$, ERDE$_{5}$, ERDE$_{50}$, and F$_{latency}$ is shown (values in bold), as well as the \emph{mean} and \emph{median} values of the results report for CLEF eRisk 2023. The second-best teams are also included.}
\label{tab:decision_T2}
\begin{tabular}{lcccccccc}
\hline
\textbf{Model}  & \textbf{P} & \textbf{R} & \textbf{F$_{1}$}    & \textbf{ERDE$_{5}$} & \textbf{ERDE$_{50}$} & \textbf{latencyTP} & \textbf{speed} & \textbf{F$_{latency}$} \\ \hline
UNSL\#0         & 0.752      & 0.767      & 0.760          & 0.048          & 0.017           & 15.0               & 0.945          & 0.718             \\
UNSL\#1         & 0.79       & 0.806      & 0.798          & 0.048          & 0.014           & 13.0               & 0.953          & 0.761             \\
UNSL\#2         & 0.752      & 0.854      & 0.800          & 0.048          & 0.013           & 14.0               & 0.949          & 0.759             \\ \hline 
ELiRF-UPV\#0    & 1.000      & 0.883      & \textbf{0.938} & \textbf{0.026} & \textbf{0.010}  & 4.0                & 0.988 & \textbf{0.927} \\
NLP-UNED-2\#1   & 0.957      & 0.883      & 0.919          & 0.034          & 0.016           & 13.0               & 0.953          & 0.876             \\
NLP-UNED-2\#4   & 0.764      & 0.883      & 0.819          & 0.033          & 0.010           & 13.0               & 0.953          & 0.781             \\ \hline 
\textit{Mean}   & 0.390      & 0.796      & 0.367          & 0.048          & 0.035           & 19.12              & 0.932          & 0.362             \\
\textit{Median} & 0.092      & 0.903      & 0.125          & 0.047          & 0.042           & 8.00               & 0.973          & 0.162             \\ \hline
\end{tabular}
\end{table}

Regarding the ranking-based metrics, the teams that achieved the best results considering 1, 100, 500, and 1000 posts were ELiRF-UPV, NLP-UNED-2, OBSER-MENH, and UNSL. As can be seen in Table \ref{tab:ranking_T2}, our team obtained the best results for the P@10 and NDCG@10 metrics. For NDCG@100, acceptable values were obtained, mainly with 100 posts. Besides, as in the decision-based metrics, UNSL\#1 and UNSL\#2 achieved similar results, outperforming UNSL\#0.

\begin{table}[h]
\caption{Ranking-based evaluation results for Task 2. Results are reported according to the three classification metrics obtained after processing 1, 100, 500, and 1000 posts, respectively.}
\label{tab:ranking_T2}
\begin{tabular}{ccccc}
\hline
\textbf{Ranking} & \textbf{Metric} & \textbf{UNSL\#0} & \textbf{UNSL\#1} & \textbf{UNSL\#2} \\ \hline
                 & P@10            & \textbf{1.00}    & \textbf{1.00}    & \textbf{1.00}    \\
1 post           & NDCG@10         & \textbf{1.00}    & \textbf{1.00}    & \textbf{1.00}    \\
                 & NDCG@100        & 0.46             & 0.57             & 0.55             \\ \hline
                 & P@10            & \textbf{1.00}    & \textbf{1.00}    & \textbf{1.00}    \\
100 posts        & NDCG@10         & \textbf{1.00}    & \textbf{1.00}    & \textbf{1.00}    \\
                 & NDCG@100        & 0.70             & 0.78             & 0.75             \\ \hline
                 & P@10            & \textbf{1.00}    & \textbf{1.00}    & \textbf{1.00}    \\
500 posts        & NDCG@10         & \textbf{1.00}    & \textbf{1.00}    & \textbf{1.00}    \\
                 & NDCG@100        & 0.64             & 0.67             & 0.69             \\ \hline
                 & P@10            & \textbf{1.00}    & \textbf{1.00}    & \textbf{1.00}    \\
1000 posts       & NDCG@10         & \textbf{1.00}    & \textbf{1.00}    & \textbf{1.00}    \\
                 & NDCG@100        & 0.64             & 0.70             & 0.69             \\ \hline
\end{tabular}
\end{table}
Finally, Table \ref{tab:teams_T2} shows the total time spent by each team to solve the task. Our team was the fastest, presenting three models with a final delay of \emph{1 day and 2 hours}, followed by the ELiRF-UPV's model (almost \emph{11 hours} of difference). The rest of the teams used five models to solve the task, with delays ranging from \emph{4 to 54 days}.
\begin{table}[h]
\caption{Total time spent by each team for Task 2. The team name, number of models, and number of user posts processed are shown. The teams are displayed according to the total time.}
\label{tab:teams_T2}
\begin{tabular}{lrrr}
\hline
\textbf{Team} & \textbf{\#models} & \textbf{\#posts processed} & \textbf{Total time}     \\ \hline
\textbf{UNSL} & \textbf{3}        & \textbf{2004}              & \textbf{1 day + 2h:17m} \\
ELiRF-UPV     & 1                 & 2004                       & 1 day + 13h:3m          \\
Xabi\_EHU     & 5                 & 2004                       & 4 days + 23h:52m        \\
OBSER-MENH    & 5                 & 2004                       & 6 days + 3h:56m         \\
RELAI         & 5                 & 764                        & 6 days + 9h:12m         \\
NLP-UNED-2    & 5                 & 2004                       & 7 days + 4h:24m         \\
NUS-eRisk     & 5                 & 2004                       & 9 days + 14h:39m        \\
BioNLP-IISERB & 5                 & 61                         & 10 days + 0h:49m        \\
SINAI         & 5               & 809                        & 10 days + 13h:0m        \\
UMUTeam       & 5                 & 2004                       & 14 days + 0h:29m        \\
NLP-UNED      & 5                 & 1151                       & 54 days + 19h:27m       \\ \hline
\end{tabular}
\end{table}
\newpage
\section{Conclusion}
\label{sec:discussion}

In this article, the UNSL team solved Tasks 1 and 2 of the eRisk 2023 Laboratory. For Task 1, we applied different approaches, obtaining better results with the proposal based on \emph{prompting}. Although we did not get the best results for this task, it would be interesting to improve these approaches considering the criteria used in the labeling process. For Task 2, we obtained outstanding results in all evaluation metrics, applying a model with extended vocabulary and a decision policy based on a historic rule. Our proposals harnessed the predictive potential of \emph{Transformers}, demonstrating that these architectures can be used in information retrieval tasks and problems of early risk detection.

%
%


\begin{thebibliography}{30}
\expandafter\ifx\csname natexlab\endcsname\relax\def\natexlab#1{#1}\fi
\providecommand{\url}[1]{\texttt{#1}}
\providecommand{\href}[2]{#2}
\providecommand{\path}[1]{#1}
\providecommand{\DOIprefix}{doi:}
\providecommand{\ArXivprefix}{arXiv:}
\providecommand{\URLprefix}{URL: }
\providecommand{\Pubmedprefix}{pmid:}
\providecommand{\doi}[1]{\href{http://dx.doi.org/#1}{\path{#1}}}
\providecommand{\Pubmed}[1]{\href{pmid:#1}{\path{#1}}}
\providecommand{\bibinfo}[2]{#2}
\ifx\xfnm\relax \def\xfnm[#1]{\unskip,\space#1}\fi
\bibitem[{Losada and Crestani(2016)}]{losada2016erisk}
\bibinfo{author}{D.~E. Losada}, \bibinfo{author}{F.~Crestani},
\newblock \bibinfo{title}{A test collection for research on depression and language use},
\newblock in: \bibinfo{booktitle}{Proc. of Conference and Labs of the Evaluation Forum (CLEF 2016)}, \bibinfo{address}{Evora, Portugal}, \bibinfo{year}{2016}, pp. \bibinfo{pages}{28--39}.
\bibitem[{Losada et~al.(2017)Losada, Crestani, and Parapar}]{losada2017erisk}
\bibinfo{author}{D.~E. Losada}, \bibinfo{author}{F.~Crestani}, \bibinfo{author}{J.~Parapar},
\newblock \bibinfo{title}{erisk 2017: Clef lab on early risk prediction on the internet: experimental foundations},
\newblock in: \bibinfo{booktitle}{International Conference of the Cross-Language Evaluation Forum for European Languages}, \bibinfo{organization}{Springer}, \bibinfo{year}{2017}, pp. \bibinfo{pages}{346--360}.
\bibitem[{Losada et~al.(2018)Losada, Crestani, and Parapar}]{losada2018overview}
\bibinfo{author}{D.~E. Losada}, \bibinfo{author}{F.~Crestani}, \bibinfo{author}{J.~Parapar},
\newblock \bibinfo{title}{Overview of erisk: early risk prediction on the internet},
\newblock in: \bibinfo{booktitle}{International Conference of the Cross-Language Evaluation Forum for European Languages}, \bibinfo{organization}{Springer}, \bibinfo{year}{2018}, pp. \bibinfo{pages}{343--361}.
\bibitem[{Losada et~al.(2019)Losada, Crestani, and Parapar}]{losada2019overview}
\bibinfo{author}{D.~E. Losada}, \bibinfo{author}{F.~Crestani}, \bibinfo{author}{J.~Parapar},
\newblock \bibinfo{title}{Overview of erisk 2019 early risk prediction on the internet},
\newblock in: \bibinfo{booktitle}{Experimental IR Meets Multilinguality, Multimodality, and Interaction: 10th International Conference of the CLEF Association, CLEF 2019, Lugano, Switzerland, September 9--12, 2019, Proceedings 10}, \bibinfo{organization}{Springer}, \bibinfo{year}{2019}, pp. \bibinfo{pages}{340--357}.
\bibitem[{Losada et~al.(2020)Losada, Crestani, and Parapar}]{losada2020erisk}
\bibinfo{author}{D.~E. Losada}, \bibinfo{author}{F.~Crestani}, \bibinfo{author}{J.~Parapar},
\newblock \bibinfo{title}{erisk 2020: Self-harm and depression challenges},
\newblock in: \bibinfo{booktitle}{Advances in Information Retrieval: 42nd European Conference on IR Research, ECIR 2020, Lisbon, Portugal, April 14--17, 2020, Proceedings, Part II 42}, \bibinfo{organization}{Springer}, \bibinfo{year}{2020}, pp. \bibinfo{pages}{557--563}.
\bibitem[{Parapar et~al.(2021)Parapar, Mart{\'\i}n-Rodilla, Losada, and Crestani}]{parapar2021overview}
\bibinfo{author}{J.~Parapar}, \bibinfo{author}{P.~Mart{\'\i}n-Rodilla}, \bibinfo{author}{D.~E. Losada}, \bibinfo{author}{F.~Crestani},
\newblock \bibinfo{title}{Overview of erisk 2021: Early risk prediction on the internet},
\newblock in: \bibinfo{booktitle}{International Conference of the Cross-Language Evaluation Forum for European Languages}, \bibinfo{organization}{Springer}, \bibinfo{year}{2021}, pp. \bibinfo{pages}{324--344}.
\bibitem[{Parapar et~al.(2022)Parapar, Mart{\'\i}n-Rodilla, Losada, and Crestani}]{parapar2022overview}
\bibinfo{author}{J.~Parapar}, \bibinfo{author}{P.~Mart{\'\i}n-Rodilla}, \bibinfo{author}{D.~E. Losada}, \bibinfo{author}{F.~Crestani},
\newblock \bibinfo{title}{Overview of erisk 2022: Early risk prediction on the internet},
\newblock in: \bibinfo{booktitle}{Experimental IR Meets Multilinguality, Multimodality, and Interaction: 13th International Conference of the CLEF Association, CLEF 2022, Bologna, Italy, September 5--8, 2022, Proceedings}, \bibinfo{organization}{Springer}, \bibinfo{year}{2022}, pp. \bibinfo{pages}{233--256}.
\bibitem[{Parapar et~al.(2023)Parapar, Mart{\'\i}n-Rodilla, Losada, and Crestani}]{parapar2023erisk}
\bibinfo{author}{J.~Parapar}, \bibinfo{author}{P.~Mart{\'\i}n-Rodilla}, \bibinfo{author}{D.~E. Losada}, \bibinfo{author}{F.~Crestani},
\newblock \bibinfo{title}{Overview of e{R}isk 2023: {E}arly {R}isk {P}rediction on the {I}nternet},
\newblock in: \bibinfo{booktitle}{Experimental IR Meets Multilinguality, Multimodality, and Interaction. 14th International Conference of the CLEF Association, CLEF 2023}, \bibinfo{organization}{Springer International Publishing, Thessaloniki, Greece}, \bibinfo{year}{2023}.
\bibitem[{Errecalde et~al.(2017)Errecalde, Villegas, Funez, Ucelay, and Cagnina}]{errecalde2017temporal}
\bibinfo{author}{M.~L. Errecalde}, \bibinfo{author}{M.~P. Villegas}, \bibinfo{author}{D.~G. Funez}, \bibinfo{author}{M.~J.~G. Ucelay}, \bibinfo{author}{L.~C. Cagnina},
\newblock \bibinfo{title}{Temporal variation of terms as concept space for early risk prediction.},
\newblock in: \bibinfo{booktitle}{Clef (working notes)}, \bibinfo{year}{2017}.
\bibitem[{Funez et~al.(2018)Funez, Ucelay, Villegas, Burdisso, Cagnina, Montes-y G{\'o}mez, and Errecalde}]{funez2018unsl}
\bibinfo{author}{D.~G. Funez}, \bibinfo{author}{M.~J.~G. Ucelay}, \bibinfo{author}{M.~P. Villegas}, \bibinfo{author}{S.~Burdisso}, \bibinfo{author}{L.~C. Cagnina}, \bibinfo{author}{M.~Montes-y G{\'o}mez}, \bibinfo{author}{M.~Errecalde},
\newblock \bibinfo{title}{Unsl's participation at erisk 2018 lab.},
\newblock in: \bibinfo{booktitle}{CLEF (Working Notes)}, \bibinfo{year}{2018}.
\bibitem[{Burdisso et~al.(2019)Burdisso, Errecalde, and Montes-y G{\'o}mez}]{burdisso2019unsl}
\bibinfo{author}{S.~G. Burdisso}, \bibinfo{author}{M.~Errecalde}, \bibinfo{author}{M.~Montes-y G{\'o}mez},
\newblock \bibinfo{title}{Unsl at erisk 2019: a unified approach for anorexia, self-harm and depression detection in social media.},
\newblock in: \bibinfo{booktitle}{CLEF (Working Notes)}, \bibinfo{year}{2019}.
\bibitem[{Loyola et~al.(2021)Loyola, Burdisso, Thompson, Cagnina, and Errecalde}]{loyola2021unsl}
\bibinfo{author}{J.~M. Loyola}, \bibinfo{author}{S.~Burdisso}, \bibinfo{author}{H.~Thompson}, \bibinfo{author}{L.~C. Cagnina}, \bibinfo{author}{M.~Errecalde},
\newblock \bibinfo{title}{Unsl at erisk 2021: A comparison of three early alert policies for early risk detection.},
\newblock in: \bibinfo{booktitle}{CLEF (Working Notes)}, \bibinfo{year}{2021}, pp. \bibinfo{pages}{992--1021}.
\bibitem[{Loyola et~al.(2022)Loyola, Thompson, Burdisso, and Errecalde}]{loyola2022unsl}
\bibinfo{author}{J.~M. Loyola}, \bibinfo{author}{H.~Thompson}, \bibinfo{author}{S.~Burdisso}, \bibinfo{author}{M.~Errecalde},
\newblock \bibinfo{title}{Unsl at erisk 2022: Decision policies with history for early classification}  (\bibinfo{year}{2022}).
\bibitem[{Vaswani et~al.(2017)Vaswani, Shazeer, Parmar, Uszkoreit, Jones, Gomez, Kaiser, and Polosukhin}]{vaswani2017attention}
\bibinfo{author}{A.~Vaswani}, \bibinfo{author}{N.~Shazeer}, \bibinfo{author}{N.~Parmar}, \bibinfo{author}{J.~Uszkoreit}, \bibinfo{author}{L.~Jones}, \bibinfo{author}{A.~N. Gomez}, \bibinfo{author}{{\L}.~Kaiser}, \bibinfo{author}{I.~Polosukhin},
\newblock \bibinfo{title}{Attention is all you need},
\newblock \bibinfo{journal}{Advances in neural information processing systems} \bibinfo{volume}{30} (\bibinfo{year}{2017}).
\bibitem[{Devlin et~al.(2018)Devlin, Chang, Lee, and Toutanova}]{devlin2018bert}
\bibinfo{author}{J.~Devlin}, \bibinfo{author}{M.-W. Chang}, \bibinfo{author}{K.~Lee}, \bibinfo{author}{K.~Toutanova},
\newblock \bibinfo{title}{Bert: Pre-training of deep bidirectional transformers for language understanding},
\newblock \bibinfo{journal}{arXiv preprint arXiv:1810.04805}  (\bibinfo{year}{2018}).
\bibitem[{Radford et~al.(2019)Radford, Wu, Child, Luan, Amodei, Sutskever et~al.}]{radford2019language}
\bibinfo{author}{A.~Radford}, \bibinfo{author}{J.~Wu}, \bibinfo{author}{R.~Child}, \bibinfo{author}{D.~Luan}, \bibinfo{author}{D.~Amodei}, \bibinfo{author}{I.~Sutskever}, et~al.,
\newblock \bibinfo{title}{Language models are unsupervised multitask learners},
\newblock \bibinfo{journal}{OpenAI blog} \bibinfo{volume}{1} (\bibinfo{year}{2019}) \bibinfo{pages}{9}.
\bibitem[{Brown et~al.(2020)Brown, Mann, Ryder, Subbiah, Kaplan, Dhariwal, Neelakantan, Shyam, Sastry, Askell et~al.}]{brown2020language}
\bibinfo{author}{T.~Brown}, \bibinfo{author}{B.~Mann}, \bibinfo{author}{N.~Ryder}, \bibinfo{author}{M.~Subbiah}, \bibinfo{author}{J.~D. Kaplan}, \bibinfo{author}{P.~Dhariwal}, \bibinfo{author}{A.~Neelakantan}, \bibinfo{author}{P.~Shyam}, \bibinfo{author}{G.~Sastry}, \bibinfo{author}{A.~Askell}, et~al.,
\newblock \bibinfo{title}{Language models are few-shot learners},
\newblock \bibinfo{journal}{Advances in neural information processing systems} \bibinfo{volume}{33} (\bibinfo{year}{2020}) \bibinfo{pages}{1877--1901}.
\bibitem[{Liu et~al.(2023)Liu, Yuan, Fu, Jiang, Hayashi, and Neubig}]{liu2023pre}
\bibinfo{author}{P.~Liu}, \bibinfo{author}{W.~Yuan}, \bibinfo{author}{J.~Fu}, \bibinfo{author}{Z.~Jiang}, \bibinfo{author}{H.~Hayashi}, \bibinfo{author}{G.~Neubig},
\newblock \bibinfo{title}{Pre-train, prompt, and predict: A systematic survey of prompting methods in natural language processing},
\newblock \bibinfo{journal}{ACM Computing Surveys} \bibinfo{volume}{55} (\bibinfo{year}{2023}) \bibinfo{pages}{1--35}.
\bibitem[{Li and Liang(2021)}]{li2021prefix}
\bibinfo{author}{X.~L. Li}, \bibinfo{author}{P.~Liang},
\newblock \bibinfo{title}{Prefix-tuning: Optimizing continuous prompts for generation},
\newblock \bibinfo{journal}{arXiv preprint arXiv:2101.00190}  (\bibinfo{year}{2021}).
\bibitem[{Liu et~al.(2019)Liu, Ott, Goyal, Du, Joshi, Chen, Levy, Lewis, Zettlemoyer, and Stoyanov}]{liu2019roberta}
\bibinfo{author}{Y.~Liu}, \bibinfo{author}{M.~Ott}, \bibinfo{author}{N.~Goyal}, \bibinfo{author}{J.~Du}, \bibinfo{author}{M.~Joshi}, \bibinfo{author}{D.~Chen}, \bibinfo{author}{O.~Levy}, \bibinfo{author}{M.~Lewis}, \bibinfo{author}{L.~Zettlemoyer}, \bibinfo{author}{V.~Stoyanov},
\newblock \bibinfo{title}{Roberta: A robustly optimized bert pretraining approach},
\newblock \bibinfo{journal}{arXiv preprint arXiv:1907.11692}  (\bibinfo{year}{2019}).
\bibitem[{OpenAI(2021)}]{openai2021}
\bibinfo{author}{OpenAI}, \bibinfo{title}{Chatgpt}, \bibinfo{howpublished}{\url{https://github.com/openai/chatgpt}}, \bibinfo{year}{2021}. \bibinfo{note}{Accessed: May 29, 2023}.
\bibitem[{Loyola et~al.(2018)Loyola, Errecalde, Escalante, and Montes~y Gomez}]{loyola2018learning}
\bibinfo{author}{J.~M. Loyola}, \bibinfo{author}{M.~L. Errecalde}, \bibinfo{author}{H.~J. Escalante}, \bibinfo{author}{M.~Montes~y Gomez},
\newblock \bibinfo{title}{Learning when to classify for early text classification},
\newblock in: \bibinfo{booktitle}{Computer Science--CACIC 2017: 23rd Argentine Congress, La Plata, Argentina, October 9-13, 2017, Revised Selected Papers 23}, \bibinfo{organization}{Springer}, \bibinfo{year}{2018}, pp. \bibinfo{pages}{24--34}.
\bibitem[{Hutto and Gilbert(2014)}]{hutto2014vader}
\bibinfo{author}{C.~Hutto}, \bibinfo{author}{E.~Gilbert},
\newblock \bibinfo{title}{Vader: A parsimonious rule-based model for sentiment analysis of social media text},
\newblock in: \bibinfo{booktitle}{Proceedings of the international AAAI conference on web and social media}, volume~\bibinfo{volume}{8}, \bibinfo{year}{2014}, pp. \bibinfo{pages}{216--225}.
\bibitem[{Lin et~al.(2017)Lin, Feng, Santos, Yu, Xiang, Zhou, and Bengio}]{lin2017structured}
\bibinfo{author}{Z.~Lin}, \bibinfo{author}{M.~Feng}, \bibinfo{author}{C.~N.~d. Santos}, \bibinfo{author}{M.~Yu}, \bibinfo{author}{B.~Xiang}, \bibinfo{author}{B.~Zhou}, \bibinfo{author}{Y.~Bengio},
\newblock \bibinfo{title}{A structured self-attentive sentence embedding},
\newblock \bibinfo{journal}{arXiv preprint arXiv:1703.03130}  (\bibinfo{year}{2017}).
\bibitem[{Mikolov et~al.(2013)Mikolov, Chen, Corrado, and Dean}]{mikolov2013efficient}
\bibinfo{author}{T.~Mikolov}, \bibinfo{author}{K.~Chen}, \bibinfo{author}{G.~Corrado}, \bibinfo{author}{J.~Dean},
\newblock \bibinfo{title}{Efficient estimation of word representations in vector space},
\newblock \bibinfo{journal}{arXiv preprint arXiv:1301.3781}  (\bibinfo{year}{2013}).
\bibitem[{Bojanowski et~al.(2017)Bojanowski, Grave, Joulin, and Mikolov}]{bojanowski2017enriching}
\bibinfo{author}{P.~Bojanowski}, \bibinfo{author}{E.~Grave}, \bibinfo{author}{A.~Joulin}, \bibinfo{author}{T.~Mikolov},
\newblock \bibinfo{title}{Enriching word vectors with subword information},
\newblock \bibinfo{journal}{Transactions of the association for computational linguistics} \bibinfo{volume}{5} (\bibinfo{year}{2017}) \bibinfo{pages}{135--146}.
\bibitem[{Petroni et~al.(2019)Petroni, Rockt{\"a}schel, Lewis, Bakhtin, Wu, Miller, and Riedel}]{petroni2019language}
\bibinfo{author}{F.~Petroni}, \bibinfo{author}{T.~Rockt{\"a}schel}, \bibinfo{author}{P.~Lewis}, \bibinfo{author}{A.~Bakhtin}, \bibinfo{author}{Y.~Wu}, \bibinfo{author}{A.~H. Miller}, \bibinfo{author}{S.~Riedel},
\newblock \bibinfo{title}{Language models as knowledge bases?},
\newblock \bibinfo{journal}{arXiv preprint arXiv:1909.01066}  (\bibinfo{year}{2019}).
\bibitem[{Qin and Eisner(2021)}]{qin2021learning}
\bibinfo{author}{G.~Qin}, \bibinfo{author}{J.~Eisner},
\newblock \bibinfo{title}{Learning how to ask: Querying lms with mixtures of soft prompts},
\newblock \bibinfo{journal}{arXiv preprint arXiv:2104.06599}  (\bibinfo{year}{2021}).
\bibitem[{Lester et~al.(2021)Lester, Al-Rfou, and Constant}]{lester2021power}
\bibinfo{author}{B.~Lester}, \bibinfo{author}{R.~Al-Rfou}, \bibinfo{author}{N.~Constant},
\newblock \bibinfo{title}{The power of scale for parameter-efficient prompt tuning},
\newblock \bibinfo{journal}{arXiv preprint arXiv:2104.08691}  (\bibinfo{year}{2021}).
\bibitem[{Burdisso et~al.(2019)Burdisso, Errecalde, and Montes-y G{\'o}mez}]{burdisso2019text}
\bibinfo{author}{S.~G. Burdisso}, \bibinfo{author}{M.~Errecalde}, \bibinfo{author}{M.~Montes-y G{\'o}mez},
\newblock \bibinfo{title}{A text classification framework for simple and effective early depression detection over social media streams},
\newblock \bibinfo{journal}{Expert Systems with Applications} \bibinfo{volume}{133} (\bibinfo{year}{2019}) \bibinfo{pages}{182--197}.

\end{thebibliography}
\end{document}